\documentclass[times,twocolumn,final,authoryear]{elsarticle}

\usepackage{prletters}
\usepackage{framed,multirow}

\RequirePackage{snapshot}

\usepackage{caption}
\usepackage{subcaption}

\usepackage{moreverb}
\usepackage{verbatim}

\usepackage{acronym}
\usepackage{enumitem}
\usepackage{color}

\usepackage{latexsym}
\usepackage{tabularx}

\usepackage{inconsolata}
\newcommand{\weblink}[1]{\texttt{#1}}

\usepackage{siunitx}
\usepackage[usenames,dvipsnames,table]{xcolor}

\usepackage{amsmath}
\usepackage{amsfonts}
\usepackage{amssymb}
\usepackage{amsthm}

\usepackage{algorithmic}
\usepackage{algorithm}
\usepackage{cases}

\usepackage{styles/mathdefs}

\renewcommand{\eqref}[1]{(\ref{eq:#1})}
\newcommand{\secref}[1]{\S\ref{sec:#1}}

\newcommand{\figref}[1]{Fig.~\ref{fig:#1}}
\newcommand{\tabref}[1]{Table~\ref{tab:#1}}

\newcommand{\avgd}{$\mathsf{avg}(\delta^*)$}
\newcommand{\stdd}{$\mathsf{std}(\delta^*)$}

\newcommand{\rc}[1]{{\color{black}#1}}

\journal{Pattern Recognition Letters}

\begin{document}


\begin{frontmatter}


\title{Automated Synchronization of Driving Data Using Vibration and Steering Events}


\author[1]{Lex \snm{Fridman}\corref{cor1}} 
\cortext[cor1]{Corresponding author: \texttt{fridman@mit.edu}}

\author[1]{Daniel E. \snm{Brown}}
\author[1]{William \snm{Angell}}
\author[1]{Irman \snm{Abdi\'c}}
\author[1]{Bryan \snm{Reimer}}
\author[2]{Hae Young \snm{Noh}}

\address[1]{Massachusetts Institute of Technology (MIT), Cambridge, MA}
\address[2]{Carnegie Mellon University (CMU), Pittsburgh, PA}


\begin{abstract}
  We propose a method for automated synchronization of vehicle sensors useful for the study of multi-modal driver
  behavior and for the design of advanced driver assistance systems. Multi-sensor decision fusion relies on synchronized
  data streams in (1) the offline supervised learning context and (2) the online prediction context. In practice, such
  data streams are often out of sync due to the absence of a real-time clock, use of multiple recording devices, or
  improper thread scheduling and data buffer management. Cross-correlation of accelerometer, telemetry, audio, and dense
  optical flow from three video sensors is used to achieve an average synchronization error of 13 milliseconds.  The
  insight underlying the effectiveness of the proposed approach is that the described sensors capture overlapping
  aspects of vehicle vibrations and vehicle steering allowing the cross-correlation function to serve as a way to
  compute the delay shift in each sensor. Furthermore, we show the decrease in synchronization error as a function of
  the duration of the data stream.
\end{abstract}



\end{frontmatter}

\section{Introduction}\label{sec:introduction}

Large multi-sensor on-road driving datasets offer the promise of helping researchers develop a better understanding of
driver behavior in the real world and aid in the design of future advanced driver assistance systems (ADAS)
\citep{fridman2016semiautomated,reimer2014driver}. As an example, the Strategic Highway Research Program (SHRP 2)
Naturalistic Driving Study includes over 3,400 drivers and vehicles with over 5,400,000 trip records
\citep{antin2011design} that contains video, telemetry, accelerometer, and other sensor data. The most interesting
insights are likely to be discovered not in the individual sensor streams but in their fusion. However, sensor fusion
requires accurate sensor synchronization. The practical challenge of fusing ``big data'', especially in the driving
domain, is that it is often poorly synchronized, especially when individual sensor streams are collected on separate
hardware \citep{meeker2014reliability}. A synchronization error of 1 second may be deemed acceptable for traditional
statistical analyses that focus on data aggregated over a multi-second or multi-minute windows. But in the driving
context, given high speed and close proximity to surrounding vehicles, a lot can happen in less than one second. We
believe that the study of behavior in relation to situationally relevant cues and the design of an ADAS system that
supports driver attention on a moment-to-moment basis requires a maximum synchronization error of 100 milliseconds. For
example, events associated with glances (e.g., eye saccades, blinks) often occur on a sub-100-millisecond timescale
\citep{mcgregor1996time}.

Hundreds of papers are written every year looking at the correlation between two or more aspects of driving (e.g., eye
movement and steering behavior). The assumption in many of these analyses is that the underlying data streams are
synchronized or aggregated over a long enough window that the synchronization error is not significantly impacting the
interpretation of the data. Often, these assumptions are not thoroughly tested. The goal of our work is to motivate the
feasibility and the importance of automated synchronization of multi-sensor driving datasets. \rc{This includes both
``one-factor synchronization'' where the passive method is the primary synchronizer and ``two-factor synchronization''
where the passive method is a validator of a real-time clock based method engineered into the data collection device.}

\rc{For the passive synchronization process, we use two event types: (1) vehicle vibration and (2) vehicle
  steering. These event types can be detected by video, audio, telemetry, and accelerometer sensors.}
Cross-correlation of processed sensor streams is used to compute the time-delay of each sensor pair. We evaluate the
automated synchronization framework on a small dataset and achieve an average synchronization error of 13
milliseconds. We also characterize the increase in accuracy with respect to increasing data stream duration which
motivates the applicability of this method to online synchronization.

The implementation tutorial and source code for this work is available at: \texttt{http://lexfridman.com/carsync}

\section{Related Work}\label{sec:related-work}

Sensor synchronization has been studied thoroughly in the domain of sensor networks where, generally, a large number of
sensor nodes are densely deployed over a geographic region to observe specific events
\citep{sivrikaya2004time,rhee2009clock}. The solution is in designing robust synchronization protocols to provide a
common notion of time to all the nodes in the sensor network \citep{elson2002fine}. These protocols rely on the ability
to propagate ground truth timing information in a master-slave or peer-to-peer framework. Our paper proposes a method
for inferring this timing information from the data itself, in a passive way as in \citep{olson2010passive}. This is only
possible when the sensors are observing largely-overlapping events. Our paper shows that up-down vibrations and
left-right turns serve as discriminating events in the driving context around which passive sensor synchronization can
be performed.

The main advantage of passive synchronization is that it requires no extra human or hardware input outside of the data
collection itself. As long as the sensors observe overlapping aspects of events in the external environment, the data
stream itself is all that is needed. The challenge of passive synchronization is that sensors capture non-overlapping
aspects of the environment as well. The overlapping aspects are the ``signal'' and the non-overlapping aspects are the
``noise''. Given this definition of signal and noise, the design of an effective passive synchronization system
requires the use of sensor pairs with low signal-to-noise ratio.

Despite its importance, very little work has been done on passive synchronization of sensors, especially in the driving
domain. This general problem was addressed in \citep{zaman2004interval} using an interval-based method for odometry and
video sensors on mobile robots. Their approach uses semi-automated and sparse event extraction. The event-extraction in
this paper is densely sampled and fully automated allowing for higher synchronization precision and greater robustness
to noisy event measurements. The pre-processing of video data for meaningful synchronizing event extraction was
performed in \citep{plotz2012automatic} for gesture recognition. We apply this idea to video data in the driving context
using dense optical flow.

Optical flow has been used in the driving domain for object detection \citep{batavia1997overtaking} and image
stabilization \citep{giachetti1998use}. Since then, dense optical flow has been successfully used for ego-motion
estimation \citep{grabe2015nonlinear}. We use the ability of optical flow to capture fast ego-motion (i.e., vibration)
for the front camera and scene vibration for the face and dash cameras in order to synchronize video data with
accelerometer data. Flow-based estimation of ego-rotation is used to synchronize front video and steering wheel
position.

\rc{
The advantages and limitations of using passive synchronization in the driving context based on vibration and steering
events can be summarized as follows:

\subsection*{Advantages:}

\begin{itemize}
\item Requires no manual human annotation before, during, or after the data collection. 
\item Requires no artificially created synchronizing events (e.g. clapping in front of the camera).
\item Requires no centralized real-time clock, and therefore does not require that the individual data streams are
  collected on the same device nor that the devices are able to communicate. So, for example, a few GoPro camcorders can
  be used to record dash, road, and face video separately and the synchronization step can then be performed after the
  videos are offloaded from the cameras.
\item Given the above three non-requirements, this approach can operate on large driving datasets that have already been
  collected in contexts where synchronization was not engineered into the hardware and software of the data collection system.
\item Can be used as validation for a centralized data collection system based on a real-time clock. If possible,
  synchronization should always be engineered into the system through both its hardware and its software before the data
  collection begins. Therefore, in a perfect world, the main application of our passive synchronization approach is for
  a second-factor validation that the final dataset is synchronized.
\end{itemize}

\subsection*{Limitations:}

\begin{itemize}
\item Requires a large amount of driving data (30+ minutes in our case) to achieve good synchronization accuracy. The
  data duration requirement may be much larger for road surfaces with low roughness indices and driving routes with
  limited amount of steering (e.g. interstate highway driving).
\item Synchronization accuracy may be affected by the positioning of the sensors and the build of the vehicle.
\item Offers no synchronization accuracy guarantees such as bounds on the performance relative to the duration of the
  driving dataset.
\item Has not been evaluated on large multi-vehicle on-road datasets that contain thousands of driven miles (e.g., SHRP
  2 \citep{antin2011design}). This is a limitation of this paper and not the approach itself. Future work that
  evaluates this approach on a larger dataset can remove this limitation and provide a more conclusive characterization
  of where this passive synchronization method succeeds and where it fails.
\end{itemize}
}

\section{Dataset and Sensors}\label{sec:sensors}

\begin{figure*}
  \centering
  \includegraphics[width=\textwidth]{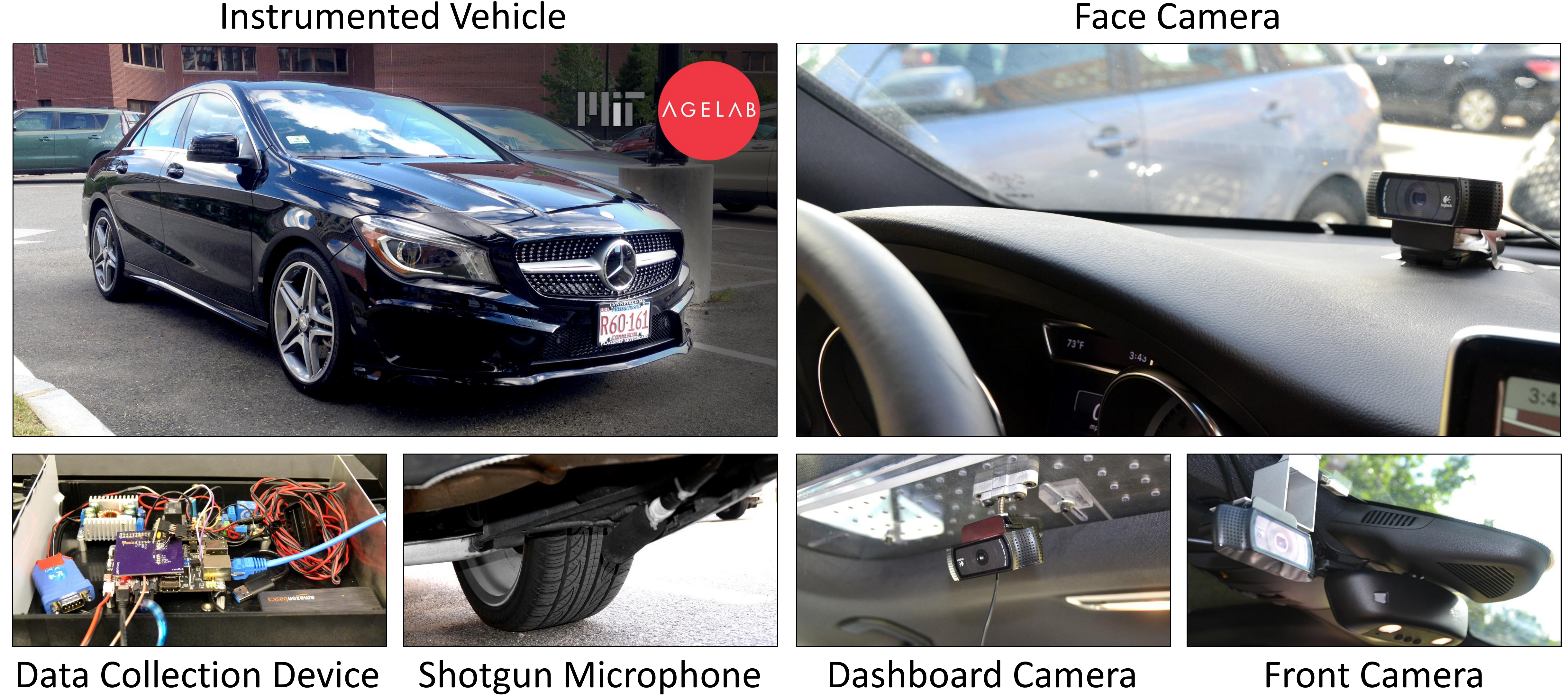}
  \caption{The instrumented vehicle, cameras, and single-board computer used for collecting the data to validate the
    proposed synchronization approach. \rc{The shotgun microphone was mounted behind the right rear tire. The face
      camera was mounted off-center of the driver's view of the roadway. The dashboard camera was mounted in the
      center of the cabin behind the driver to include a view of the instrument cluster, center stack, and the driver's
      body. The forward roadway camera was mounted to the right of the rearview mirror as close to the center line of
      the vehicle as possible (i.e. next to the OEM interior upper windshield enclosure). The IMU sensor was on the data
      collection device that was placed behind the driver's seat.}}
  \label{fig:car}
\end{figure*}
 
\begin{figure*}
  \centering
  \includegraphics[width=\textwidth]{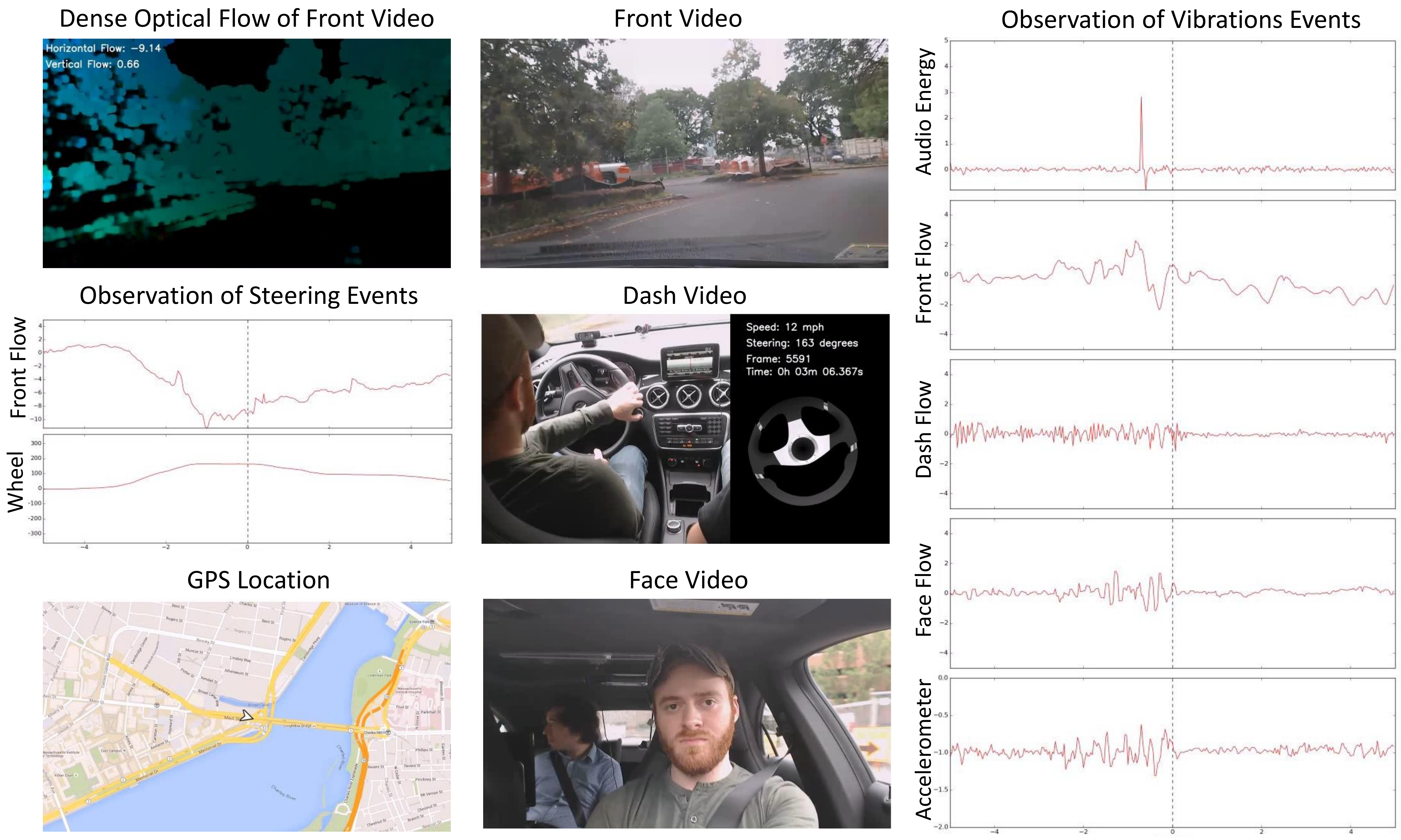}
  \caption{Snapshot from a 30 fps video visualization of a synchronized set of sensors collected for the 37 minute
    experimental run used throughout the paper as an example. \rc{The plots show a 10 second window around the current
    moment captured by the three cameras, the dense optical flow of the front video, and the GPS location on a map in
    the other subfigures.} Full video is available online at: \weblink{http://lexfridman.com/carsync}}
  \label{fig:carsync-screenshot}
\end{figure*}

In order to validate the proposed synchronization approach we instrumented a 2014 Mercedes CLA with a single-board
computer and consumer-level inexpensive sensors: 3 webcams, a shotgun microphone behind the rear tire, GPS, an IMU
module, and a CAN controller for vehicle telemetry. \rc{The instrumented vehicle is shown in \figref{car}. Details on
  the positioning of the sensors are provided in the figure's caption. Through empirical testing we found that small
  changes in the position of the sensors did not have any noticeable impact on synchronization accuracy.}

The collection of data was performed on 5 runs, each time traveling the same route in different traffic conditions. The
duration of each run spanned from 37 minutes to 68 minutes. Unless otherwise noted, the illustrative figures in this
paper are based on the 37 minute run.

Three manual synchronization techniques were used on each run to ensure that perfect synchronization was achieved and
thus can serve as the ground truth for the proposed automated synchronization framework:

\begin{enumerate}
\item The same millisecond-resolution clock was placed in front of each camera at the beginning and end of each
  run. This allowed us to manually synchronize the videos together.
\item We clapped three times at the beginning and the end of the each run. This was done in front of the camera such
  that the tire microphone could pick up the sound of each clap. This allowed us to manually synchronize the audio and
  the video.
\item We visualized the steering wheel (see \figref{carsync-screenshot}) according to the steering wheel position
  reported in the CAN and lined it up to the steering wheel position in the video of the dashboard. This allows us to
  manually synchronize video and telemetry.
\end{enumerate}

A real-time clock module was used to assign timestamps to all discrete samples of sensor data. This timestamp and the
above three manual synchronization methods were used to produce the ground truth dataset over which the evaluation in
\secref{synchronization} is performed.

\subsection{Sensors}

The following separate sensor streams are collected and synchronized in this work:

\begin{itemize}
\item \textbf{Front Video Camera}: 720p 30fps video of the forward roadway. Most of the optical flow motion in the video is of the
  external environment. Therefore, vibration is captured through ego-motion estimated by the vertical component of the
  optical flow. Steering events are captured through the horizontal component of the optical flow. See \secref{optical-flow}.
\item \textbf{Dash Video Camera}: 720p 30fps video of the dashboard. This is the most static of the video streams, so spatially-averaged
  optical flow provides the most accurate estimate of vibrations.
\item \textbf{Face Video Camera}: 720p 30fps video of the driver's face. This video stream is similar to dashboard video except
  for the movements of the driver. These movements contribute noise to the optical flow vibration estimate.
\item \textbf{Inertial Measurement Unit (IMU)}: Accelerometer used to capture the up-down vibrations of the vehicle that
  correspond to y-axis vibrations in the video. Average sample rate is 48 Hz.
\item \textbf{Audio}: Shotgun microphone attached behind the right rear tire of the vehicle used to capture the interaction
  of the tire with the surface. Sample rate is 44,100 Hz and bit depth is 16.
\item \textbf{Vehicle Telemetry}: Parsed messages from the controller area network (CAN) vehicle bus reduced down in this work to
  just steering wheel position. Sample rate is 100Hz.
\end{itemize}

A snapshot from a video visualization of these sensor streams is shown in \figref{carsync-screenshot}. GPS position was
collected but not used as part of the synchronization because its average sample rate is 1 Hz which is 1 to 2 orders of
magnitude less frequent than the other sensors.

\subsection{Dense Optical Flow}\label{sec:optical-flow}

The dense optical flow is computed as a function of two images taken at times $t$ and $t + \Delta t$, where $\Delta t$
varies according to the frame rate of the video (30 fps in the case of this paper) and the burstiness of frames due to
the video buffer size. We use the Farneback algorithm \citep{farneback2003two} to compute the dense optical flow. It
estimates the displacement field using a quadratic polynomial for the neighborhood of each pixel. The algorithm assumes
a slowly varying displacement field, which is a valid assumption for the application of detecting vibrations and
steering since those are events which affect the whole image uniformly relative to the depth of the object in the
image. Segmentation was used to remove non-static objects, but it did not significantly affect the total magnitude of
either the $x$ or $y$ components of the flow. The resulting algorithm produces a flow for which the following holds:

\begin{equation}
I(x,y,t) = I(x+\Delta x, y + \Delta y, t + \Delta t)
\end{equation}

\noindent where $I(x, y, t)$ is the intensity of the pixel $(x, y)$ at time $t$, $\Delta x$ is the $x$ (horizontal)
component of the flow, and $\Delta y$ is the $y$ (vertical) component of the flow. These two components are used
separately as part of the synchronization. The vertical component is used to measure the vibration of the vehicle and the
horizontal component is used to measure the turning of the vehicle.

\section{Synchronization Framework}\label{sec:synchronization}

Unless otherwise noted, the figures in this section show sensor traces and cross correlation functions for a single 37
minute example run. The two synchronizing event types are vibrations and steering, both densely represented throughout a
typical driving session.

\begin{figure*}
  \centering
  \begin{subfigure}[t]{\textwidth}
    \includegraphics[width=\textwidth]{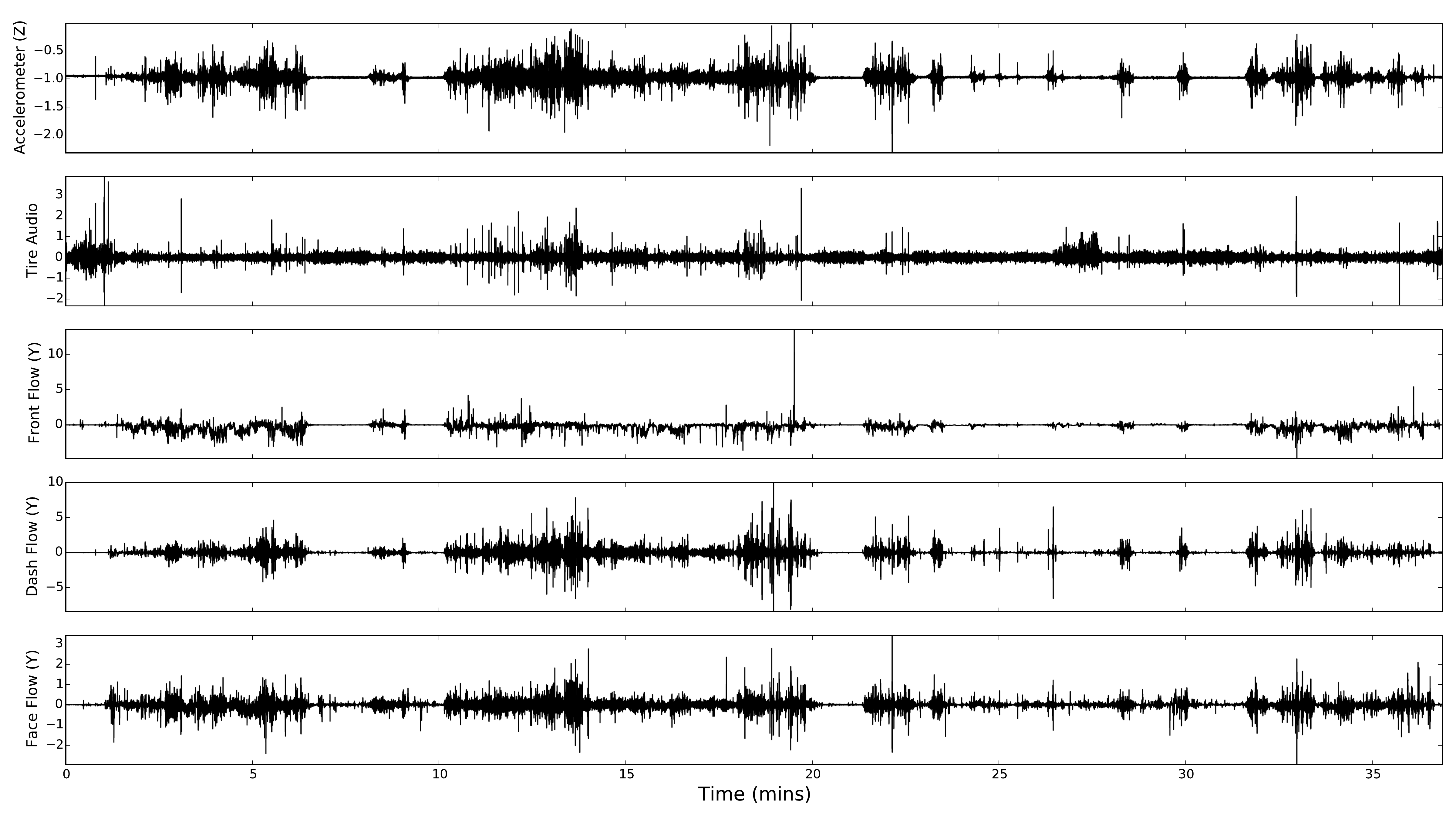}
    \caption{The accelerometer, video, and audio sensors capturing the vibration of the vehicle. The x-axis is time in
      minutes and the y-axis is the value of the sensor reading.}
  \end{subfigure}\\\vspace{0.4in}
  \begin{subfigure}[t]{\textwidth}
    \includegraphics[width=\textwidth]{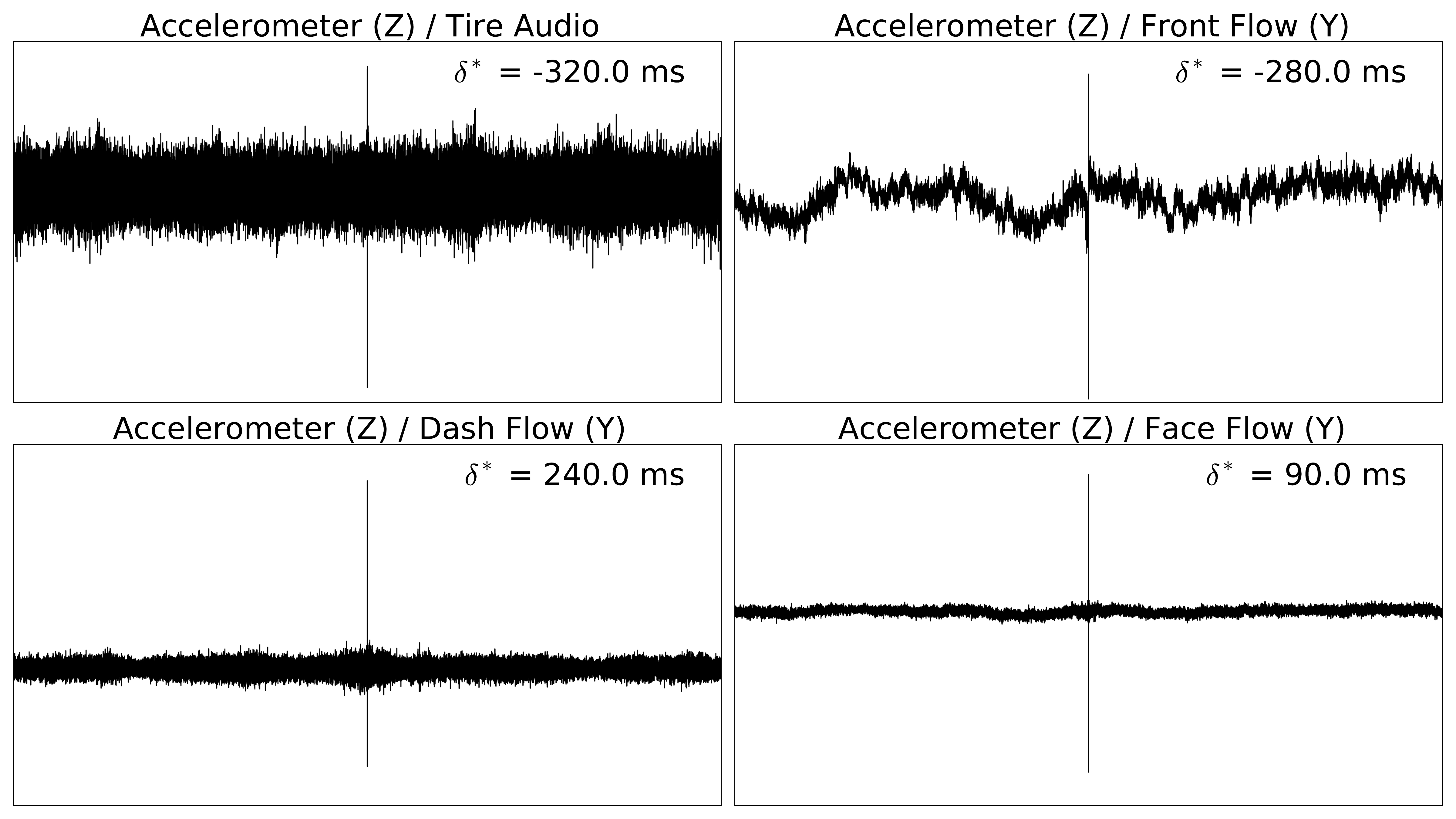}
    \caption{The cross-correlation functions and optimal time-delay $\delta^*$ of audio and video with respect to the
      accelerometer. The x-axis is the shift $t$ in \eqref{cross-correlation} and the y-axis is the magnitude of the
      correlation.}
    \label{fig:vibration-function}
  \end{subfigure}
  \caption{The vibration-based synchronization for the 37-minute example run.}
  \label{fig:vibration-results}
\end{figure*}

\subsection{Cross-Correlation}\label{sec:cross-correlation}

Cross-correlation has long been used as a way to estimate time delay between two regularly sampled signals
\citep{knapp1976generalized}. We use an efficient FFT-based approach for computing the cross correlation function
\citep{lewis1995fast} with an $O(n \log n)$ running time complexity (compared to $O(n^2)$ running time of the naive implementation):

\begin{equation}\label{eq:cross-correlation}
(f \star g)[t]\ \stackrel{\mathrm{def}}{=} \sum_{i=-\infty}^{\infty} f[i]\ g[i+t]
\end{equation}

\noindent where $f$ and $g$ are real-valued discrete functions, $t$ is the time shift of $g$, and both $f$ and $g$ are
zero for $i$ and $i+t$ outside the domain of $f$ and $g$ respectively.

The optimal time shift $\delta^*$ for synchronizing the two data streams is computed by choosing the $t$ that maximizes
the cross correlation function:

\begin{equation}\label{eq:optimal-delay}
\delta^*=\underset{t}{\operatorname{arg\,max}}(f \star g)[t]
\end{equation}

This optimization assumes that the optimal shift corresponds to maximum positive correlation. Three of the sensors under
consideration are negatively correlated: (1) vertical component of face video optical flow, (2) vertical component of
dash video optical flow, and (3) horizontal component of front video optical flow. These three were multiplied by -1, so
that all sensors used for synchronization are positively correlated.

Vibration and steering events are present in all vehicle sensors, but post processing is required to articulate these
events in the data. Accelerometer and steering wheel position require no post-processing for the cross correlation
computation in \eqref{optimal-delay}. The three video streams were processed to extract horizontal and vertical
components of dense optical flow as discussed in \secref{optical-flow}. The shotgun microphone audio was processed by
summing the audio energy in each 10 ms increment. Several filtering methods (Wiener filter, total variation denoising,
and stationary wavelet transform) under various parameter settings were explored programmatically, but they did not improve
the optimal time shift computation accuracy as compared to cross correlation of un-filtered sensor data.

\figref{vibration-results} shows the sensor trace and cross correlation functions for the z component of acceleration,
the y component of dense optical flow for the three videos, and discretized audio energy. These data streams measure the
vibration of the vehicle during the driving session. All of them capture major road bumps and potholes. The audio captures
more complex properties of the surface which makes vibration-based synchronization with audio the most noisy of the five
sensors. These 5 sensors can be paired in 10 ways. We found that the most robust and least noisy cross correlation
function optimization was for the pairing all sensors with the accelerometer. This is intuitive since the accelerometer
is best able to capture vibration. The resulting 4 cross correlation functions for the 37 minute example run are shown
in \figref{vibration-function}.

\begin{figure}
  \centering
  \begin{subfigure}[t]{\columnwidth}
    \includegraphics[width=\textwidth]{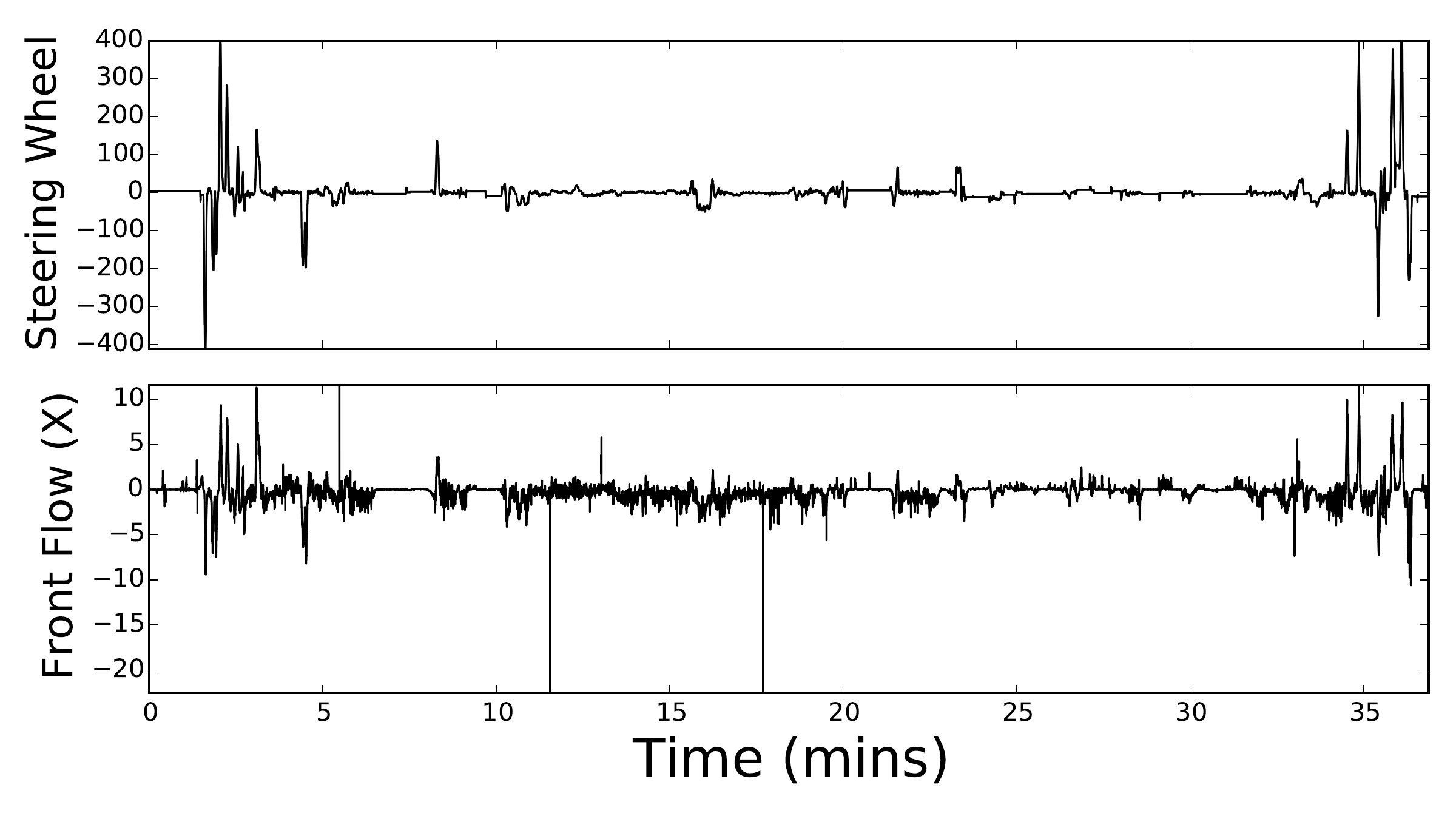}
    \caption{The telemetry and video sensors capturing the steering of the vehicle. The x-axis is time in
      minutes and the y-axis is the value of the sensor reading.}
  \end{subfigure}
  \begin{subfigure}[t]{\columnwidth}
    \includegraphics[width=\textwidth]{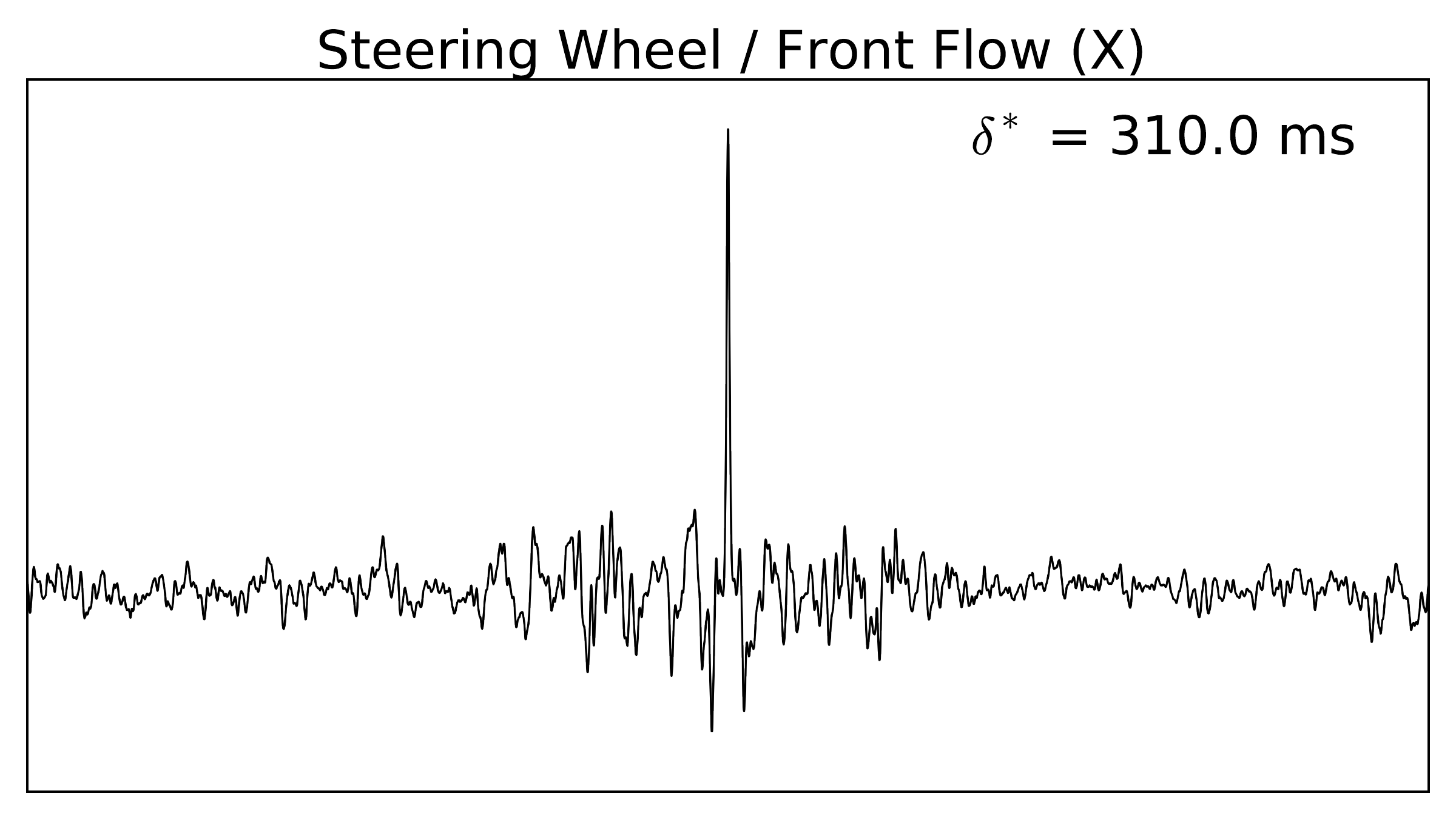}
    \caption{The cross-correlation functions and optimal time-delay $\delta^*$ of telemetry and front video. The x-axis
      is the shift $t$ in \eqref{cross-correlation} and the y-axis is the magnitude of the correlation.}
  \end{subfigure}
  \caption{The steering-based synchronization for the 37-minute example run.}
  \label{fig:turning-results}
\end{figure}

\figref{turning-results} shows the sensor trace and cross correlation function for the x component of dense optical flow
for the front video and the position of the steering wheel. This is an intuitive pairing of sensors that produced
accurate results for our experiments in a single vehicle. However, since the way a vehicle's movement corresponds to
steering wheel position depends on the sensitivity of the steering wheel, the normalizing sensor-pair delay (see
\secref{online-synchronization}) may vary from vehicle to vehicle.

\subsection{Online and Offline Synchronization}\label{sec:online-synchronization}

\definecolor{lightHyo}{gray}{0.7}
\newcommand{\headerHyo}[1]{\rule[-1.2em]{0em}{3em}\renewcommand{\arraystretch}{1}\begin{tabular}[c]{@{}l@{}}#1\end{tabular}}
\renewcommand{\arraystretch}{1.5}
\begin{table}[h!]
  \centering
  \caption{The mean and standard deviation of the optimal delay $\delta^*$ in \eqref{optimal-delay}. The mean serves as
    the normalizing sensor-pair delay. The standard
    deviation, in this case, is an estimate for average synchronization error. Across the five sensor pairs listed
    here, the average error is 13.5 ms.}
  \begin{tabular}{lll}
    \hline
    \headerHyo{Sensor Pair} &
    \headerHyo{\avgd\\(ms)}&
    \headerHyo{\stdd\\(ms)}\\
    \hline
    Accelerometer / Tire Audio& 305.2 & 22.8\\\arrayrulecolor{lightHyo}\hline
    Accelerometer / Front Flow& -279.9 & 12.7\\\arrayrulecolor{lightHyo}\hline
    Accelerometer / Dash Flow& 246.4 & 8.7\\\arrayrulecolor{lightHyo}\hline
    Accelerometer / Face Flow& 95.0 & 14.2\\\arrayrulecolor{lightHyo}\hline
    Steering Wheel / Front Flow& 312.1 & 9.3\\\arrayrulecolor{black}\hline
  \end{tabular}
  \label{tab:sync-performance}
\end{table}

\begin{figure}
  \centering
  \includegraphics[width=\columnwidth]{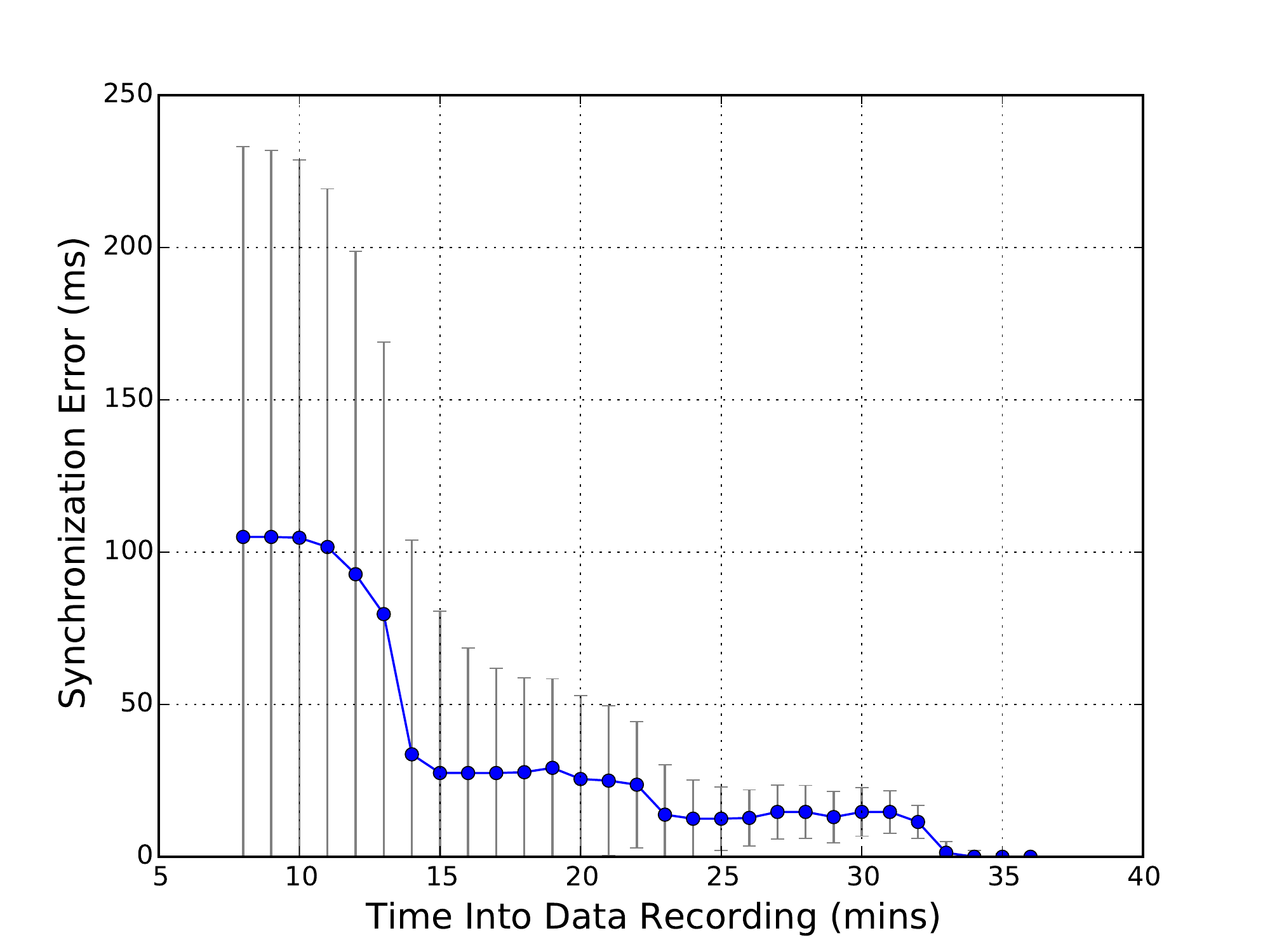}
  \caption{The decrease of synchronization error versus the duration of the data stream. The mean and standard deviation
    forming the points and errorbars in the plot are computed over 5 sensor pairs and over 5 runs, each of which
    involved the instrumented vehicle traveling same route (lasting 37-68 minutes).}
  \label{fig:online}
\end{figure}

\tabref{sync-performance} shows the results of computing the optimal delay $\delta^*$ for each sensor pairing in each of
the 5 runs as discussed in \secref{cross-correlation}. The value \avgd for each sensor pairing is the ``normalizing
delay'', which is an estimate of the delay inherent in the fact that optical flow, audio energy, accelerometer,
and steering wheel position are capturing different temporal characteristics of the same events. For example, there is a
consistent delay of just over 300ms between steering wheel position and horizontal optical flow in the front camera. 
This normalizing delay is to be subtracted from $\delta^*$ computed on future data in order to determine the best time shift
for synchronizing the pair of sensor streams. In this case, the standard deviation \stdd is an estimate of the
synchronization error. For the 5 runs in our dataset, the average error is 13.5 ms which satisfies the goal of sub-100 ms
accuracy stated in \secref{introduction}.

The proposed synchronization framework is designed as an offline system for post-processing sensor data after the data
collection has stopped. However, we also consider the tradeoff between data stream duration and synchronization accuracy
in order to evaluate the feasibility of this kind of passive synchronization to be used in an online real-time
system. \figref{online} shows the decrease in synchronization error versus the duration of the data stream. Each point
averages 5 sensor pairings over 5 runs. While the duration of each run ranged from 37 to 68 minutes, for this plot we only average
over the first 37 minutes of each run. The synchronization error here is a measurement of the difference between the
current estimate of $\delta^*$ and the one converged to after the full sample is considered. This error does not
consider the ground truth which is estimated to be within 13.5 ms of this value. Data streams of duration less than 8
minutes produced synchronization errors 1-2 orders of magnitude higher than the ones in this plot. The takeaway from
this tradeoff plot is that an online system requires 10 minutes of data to synchronize the multi-sensor stream to a degree
that allows it to make real-time decisions based on the fusion of these sensors. 

\section{Conclusion}\label{sec:conclusion}

Analysis and prediction based on fusion of multi-sensor driving data requires that the data is synchronized. We propose
a method for automated synchronization of vehicle sensors based on vibration and steering events. This approach is
applicable in both an offline context (i.e., for driver behavior analysis) and an online context (i.e., for real-time
intelligent driver assistance). We show that a synchronization error of 13.5 ms can be achieved for a driving session of
35 minutes.

\section*{Acknowledgment}

Support for this work was provided by the New England University Transportation Center, and the Toyota Class Action
Settlement Safety Research and Education Program. The views and conclusions being expressed are those of the authors,
and have not been sponsored, approved, or endorsed by Toyota or plaintiffs’ class counsel.

\bibliographystyle{model2-names}
\bibliography{sync,lex-fridman,agelab}

\end{document}